\RequirePackage{fix-cm}
\documentclass[onecolumn,10pt]{svjour3}
\smartqed  
\usepackage{graphicx}
\usepackage{fixltx2e}
\usepackage[misc]{ifsym}
\usepackage{float}
\usepackage[multi-part-units=single]{siunitx}
\usepackage[font=small,labelfont=bf,tableposition=top]{caption}

\DeclareCaptionLabelFormat{andtable}{#1~#2  \&  \tablename~\thetable}
\begin{document}

\title{Plaque Classification in Coronary Arteries from IVOCT Images Using Convolutional Neural Networks and Transfer Learning}


\titlerunning{Plaque Classification in Coronary Arteries}        

\author{Nils Gessert$^1$        \and
        Markus Heyder$^1$  \and
        Sarah Latus$^1$  \and
        Matthias Lutz$^2$ \and
        Alexander Schlaefer$^1$  
}


\institute{\Letter \quad Nils Gessert, \email{nils.gessert@tuhh.de}, Tel.: +49 (0)40 42878 3389, https://orcid.org/0000-0001-6325-5092 \\ \\ $^1$ Hamburg University of Technology, Schwarzenbergstra\ss{}e 95, 21073 Hamburg \\ $^2$ Universitätsklinikum Schleswig-Holstein, Arnold-Heller-Straße 3, 24105 Kiel}

\date{Preprint, submitted to CARS 2018, accepted for publication}

\maketitle

\begin{abstract}

Advanced atherosclerosis in the coronary arteries is one of the leading causes of deaths worldwide while being preventable and treatable. In order to image atherosclerotic lesions (plaque), intravascular optical coherence tomography (IVOCT) can be used. The technique provides high-resolution images of arterial walls which allows for early plaque detection by experts. Due to the vast amount of IVOCT images acquired in clinical routines, automatic plaque detection has been addressed. For example, attenuation profiles in single A-Scans of IVOCT images are examined to detect plaque. We address automatic plaque classification from entire IVOCT images, the cross-sectional view of the artery, using deep feature learning. In this way, we take context between A-Scans into account and we directly learn relevant features from the image source without the need for handcrafting features.

\keywords{IVOCT \and Plaque Classifcation \and Deep Learning \and CNN}
\end{abstract}

\section{Methods} 

\begin{figure*}
  \centering
  \includegraphics[width=1\textwidth]{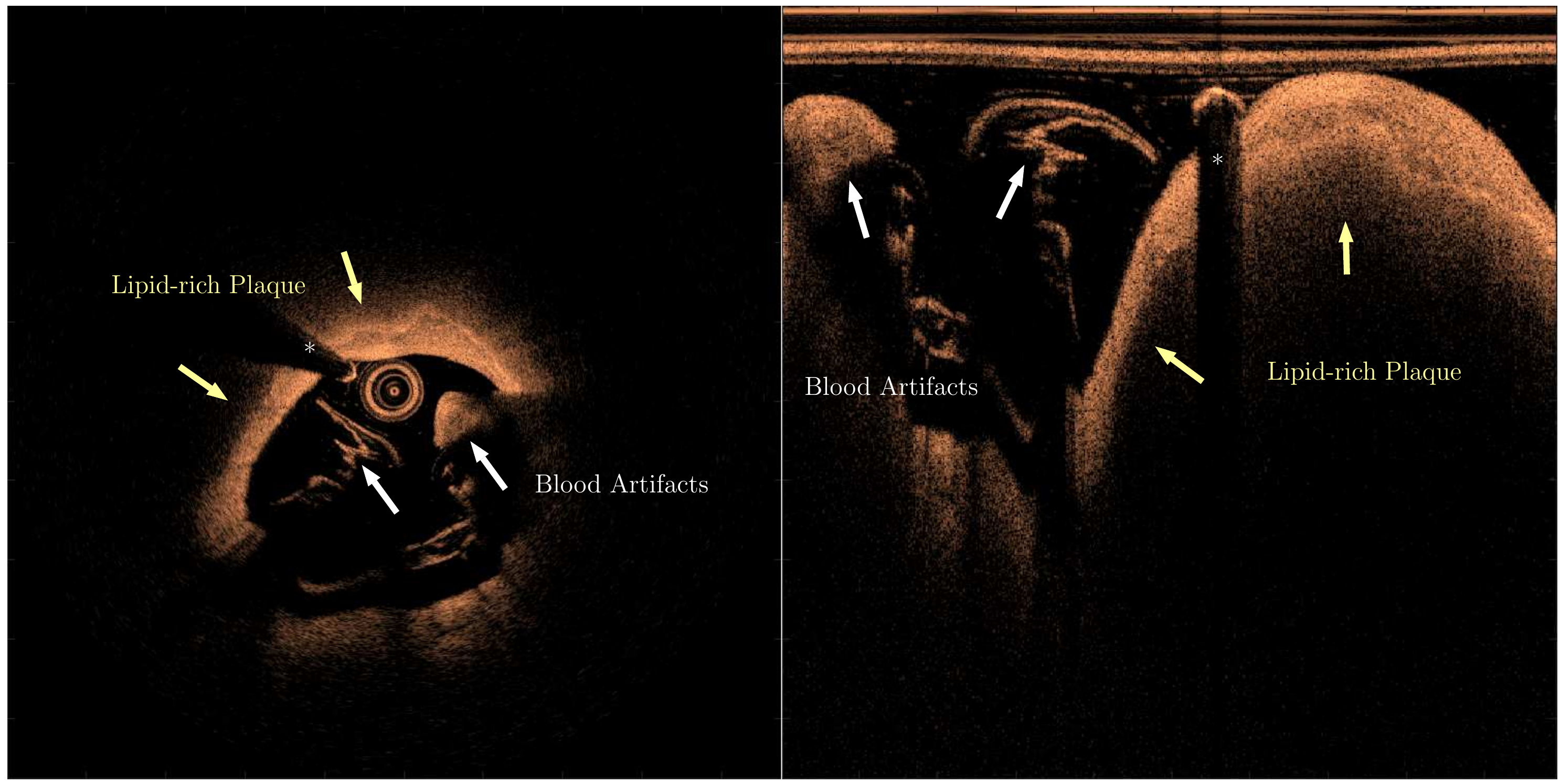}
\caption{Left, a cartesian IVOCT image is shown. Right, the polar image is shown. Note, that there are various artifacts in the vessel which impede a clear view of the vessel wall. Lipid-rich plaque is visible in this image. ‘*’ denotes the guide-wire artifact.}
\label{fig:ivoct}       
\end{figure*}

We built a new database of IVOCT images using in-vivo patient data acquired with a St. Jude Medical Ilumien OPTIS. A trained expert with daily application experience with IVOCT provides the ground-truth labels for the images. Each B-Scan is assigned a binary label of “plaque” or “no plaque”. In total, the dataset contains 41 patients with 2868 labeled B-Scans. We split off a test set of 6 patients with 509 B-Scans. The dataset is slightly imbalanced with \SI{45}{\percent} of the B-Scans being labelled as “no plaque”. 
In contrast to previous approaches \cite{ughi2013automated,rico2016automatic}, we do not apply extensive pre-processing for lumen segmentation and flattening or guide-wire artifact and catheter removal. Therefore, we force our models to be robust towards all kinds of artifacts which will often appear in clinical practice. This allows our models to deal with any raw data without having to rely on an automatic segmentation procedure which might fail if the artery wall is not consistently visible, as shown in Figure\ref{fig:ivoct}.
We employ convolutional neural networks (CNNs) to directly learn relevant features for plaque classification from the B-Scans.  We make use of standard architectures for image classification and object detection in non-medical settings. The architectures are Resnet50, Resnet101, InceptionV3 and Inception-ResnetV2 \cite{litjens2017survey,Szegedy.2017}. Moreover, we apply transfer learning which can help with the adaptation to new problem domains where data is limited \cite{ravishankar2016understanding}. Therefore, we pretrain the models on the ImageNet dataset which contains 1.2 million natural images with 1000 classes. We remove the models’ last layer and add a layer with two outputs for binary classification. In their original design, the Inception-based models employ dropout before the output. For comparability, we employ dropout with a probability of p=0.5 before the output of every model.
The images that are fed into the model can be represented either in polar or cartesian form, see Figure~\ref{fig:ivoct}. In polar form, the acquired A-Scans are aligned next to each other in temporal acquisition order. Each A-Scan represents a single depth scan at a certain angle of the artery cross-section. The polar image can be transformed into cartesian space by mapping the A-Scans to their respective angle and applying interpolation in between. This representation provides a more intuitive cross-sectional view of the artery and is therefore used by practitioners. From an image processing point of view, both should capture the same amount of information. We investigate whether either representation is more advantageous for deep feature learning.
We resize the input images to a size of 300x300 pixels. For data augmentation, we apply random cropping with a patch size of 270x270 pixels during training. Furthermore, we apply random rotations to the cartesian images and we randomly flip polar images along the temporal axis. For evaluation, we use a single center crop of the training patch size without flipping or rotations.

\section{Results}

\begin{figure*}
  \centering
  \includegraphics[width=1\textwidth]{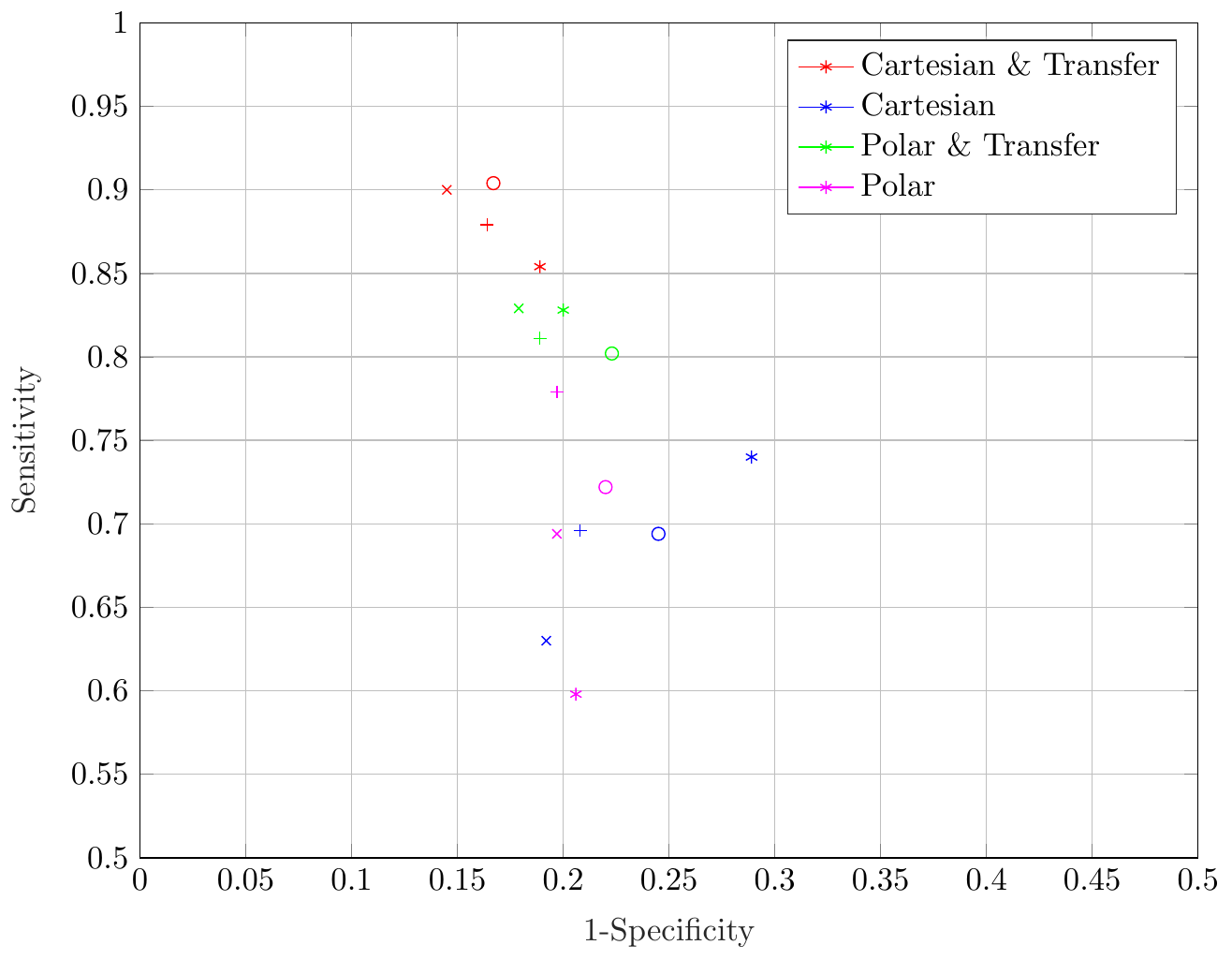}
\caption{The sensitivity and 1-specificity of 16 different models for binary plaque classification is shown. Each mark represents one network architecture. ‘*’ refers to ResNet50, ‘x’ refers to ResNet101, ‘o’ refers to InceptionV3 and ‘+’ refers to Inception-ResnetV2.}
\label{fig:res}       
\end{figure*}

The resulting prediction performance of our models on the test set is shown in Figure~\ref{fig:res}. The sensitivity and 1-specificity of each model for classification of an image as “plaque” is shown. The four models were each trained on polar and cartesian images, both with and without transfer learning. Models in the upper left corner perform better as they have a higher sensitivity and specificity. 
Overall, pretraining on ImageNet appears to improve performance significantly as the best model without pretraining shows an overall accuracy of \SI{78.9}{\percent} (Inception-ResnetV2 with polar images) with a sensitivity of \SI{77.9}{\percent} and a specificity of \SI{80.3}{\percent} while the best model with pretraining (ResNet101 with cartesian images) shows an overall accuracy of  \SI{88.0}{\percent} with a sensitivity of \SI{90.0}{\percent} and a specificity of \SI{85.5}{\percent}. It appears, that meaningful feature transfer from the natural image domain to the IVOCT image domain was achieved. 
Also, using cartesian representations results in better classification performance. All models with pretraining achieve both a higher sensitivity and specificity when being trained on cartesian images. For example, the best model with polar images shows an accuracy of \SI{82.8}{\percent} compared to \SI{88.0}{\percent} for cartesian images (both Resnet101). This indicates that a cartesian real-world image representation helps CNN-based learning when employing transfer learning. Without pretraining, the difference is not as clear and some models perform better with polar representations.
The different models all perform similar with Resnet101 standing out slightly as it performs best for the two pretrained cases. All in all, the choice of image representation and transfer learning has a larger impact on performance than the model choice.

\section{Conclusions}

We perform plaque classification from IVOCT images using CNNs for deep feature learning. For this purpose, we build a database with in-vivo patient image data that is labelled by a trained expert. We allow images with various artifacts in our dataset und force our models to learn robustness. We employ various standard CNN models with additional pretraining on ImageNet for transfer learning. Our results show that pretraining significantly boosts performance. Moreover, using cartesian image representations appears to be beneficial for CNN learning. Overall, our best model achieves an accuracy of \SI{88.0}{\percent} for plaque classification.


\bibliographystyle{spmpsci} 
\bibliography{egbib}   

\begin{thebibliography}{1}
\providecommand{\url}[1]{{#1}}
\providecommand{\urlprefix}{URL }
\expandafter\ifx\csname urlstyle\endcsname\relax
  \providecommand{\doi}[1]{DOI~\discretionary{}{}{}#1}\else
  \providecommand{\doi}{DOI~\discretionary{}{}{}\begingroup
  \urlstyle{rm}\Url}\fi

\bibitem{ughi2013automated}
Ughi, G.J., Adriaenssens, T., Sinnaeve, P., Desmet, W., D’hooge, J. (2013)
  Automated tissue characterization of in vivo atherosclerotic plaques by
  intravascular optical coherence tomography images.
\newblock Biomedical optics express \textbf{4}(7), 1014--1030

\bibitem{rico2016automatic}
Rico-Jimenez, J.J., Campos-Delgado, D.U., Villiger, M., Otsuka, K., Bouma,
  B.E., Jo, J.A. (2016) Automatic classification of atherosclerotic plaques
  imaged with intravascular oct.
\newblock Biomedical optics express \textbf{7}(10), 4069--4085

\bibitem{litjens2017survey}
Litjens, G., Kooi, T., Bejnordi, B.E., Setio, A.A.A., Ciompi, F., Ghafoorian,
  M., van~der Laak, J.A., van Ginneken, B., S{\'a}nchez, C.I. (2017) A survey
  on deep learning in medical image analysis.
\newblock Medical image analysis \textbf{42}, 60--88

\bibitem{Szegedy.2017}
Szegedy, C., Ioffe, S., Vanhoucke, V., Alemi, A.A. (2017) {Inception-v4,
  Inception-ResNet and the Impact of Residual Connections on Learning}.
\newblock In: {AAAI}, pp. 4278--4284

\bibitem{ravishankar2016understanding}
Ravishankar, H., Sudhakar, P., Venkataramani, R., Thiruvenkadam, S., Annangi,
  P., Babu, N., Vaidya, V. (2016) Understanding the mechanisms of deep transfer
  learning for medical images.
\newblock In: Deep Learning and Data Labeling for Medical Applications, pp.
  188--196. Springer

\end{thebibliography}

\end{document}